\documentclass{article}

\usepackage[final]{neurips_2024}
\usepackage{graphicx}
\usepackage{booktabs}
\usepackage{multirow}
\usepackage{multicol}
\usepackage{caption}
\usepackage{mathtools}
\usepackage{adjustbox}
\usepackage{colortbl}
\usepackage[accsupp]{axessibility}  % Improves PDF readability 

\usepackage[utf8]{inputenc} % allow utf-8 input
\usepackage[T1]{fontenc}    % use 8-bit T1 fonts
\usepackage{hyperref}       % hyperlinks
\usepackage{url}            % simple URL typesetting
\usepackage{booktabs}       % professional-quality tables
\usepackage{amsfonts}       % blackboard math symbols
\usepackage{nicefrac}       % compact symbols for 1/2, etc.
\usepackage{microtype}      % microtypography
\usepackage{xcolor}         % colors

\title{You Only Look Around: Learning Illumination Invariant Feature for Low-light Object Detection}

\author{%
  Mingbo Hong\thanks{Equal contribution.}\\
  Megvii Technology, Beijing, China\\
  \texttt{mingbohong97@gmail.com} \\
\And
Shen Cheng$^{*}$\\
Megvii Technology, Beijing, China\\
\texttt{chengshen@megvii.com} \\
\And
Haibin Huang \\
Kuaishou Technology\\
\texttt{jackiehuanghaibin@gmail.com} \\
\And
Haoqiang Fan \\
Megvii Technology, Beijing, China\\
\texttt{fhq@megvii.com} \\
\And
Shuaicheng Liu\thanks{Corresponding Author.} \\
University of Electronic Science and Technology of China\\
\texttt{liushuaicheng@uestc.edu.cn} \\
}

\begin{document}

\maketitle

\begin{abstract}
% In this paper, we introduce YOLA, a novel framework for object detection in low-light scenarios. Unlike previous works, we propose to tackle this challenging problem from the perspective of feature learning. Specifically, we propose to learn illumination-invariant features through the Lambertian image formation model. We observe that, under the Lambertian assumption, it is feasible to approximate illumination-invariant feature maps by exploiting the interrelationships between neighboring color channels and spatially adjacent pixels. By incorporating additional constraints, these relationships can be characterized in the form of convolutional kernels, making them trainable within a network. Consequently, we introduce a novel module dedicated to the extraction of illumination-invariant features from low-light images, complemented by a newly designed loss function that reinforces feature consistency across varying illumination levels. Both modules can be easily integrated into existing object detection frameworks, and our empirical results demonstrate a notable enhancement in performance for low-light object detection tasks. 
In this paper, we introduce YOLA, a novel framework for object detection in low-light scenarios. Unlike previous works, we propose to tackle this challenging problem from the perspective of feature learning. Specifically, we propose to learn illumination-invariant features through the Lambertian image formation model. We observe that, under the Lambertian assumption, it is feasible to approximate illumination-invariant feature maps by exploiting the interrelationships between neighboring color channels and spatially adjacent pixels. By incorporating additional constraints, these relationships can be characterized in the form of convolutional kernels, which can be trained in a detection-driven manner within a network. Towards this end, we introduce a novel module dedicated to the extraction of illumination-invariant features from low-light images, which can be easily integrated into existing object detection frameworks. Our empirical findings reveal significant improvements in low-light object detection tasks, as well as promising results in both well-lit and over-lit scenarios. Code is available at
\url{https://github.com/MingboHong/YOLA}.
% In this paper, we introduce YOLA, a novel framework for object detection in low-light scenarios. Unlike previous works, we propose to tackle this challenging problem from the perspective of feature learning. Specifically, we propose to learn illumination-invariant features through the Lambertian image formation model. We observe that, under the Lambertian assumption, it is feasible to approximate illumination-invariant feature maps by exploiting the interrelationships between neighboring color channels and spatially adjacent pixels. By incorporating additional constraints, these relationships can be characterized in the form of convolutional kernels, making them trainable within a network. Consequently, we introduce a novel module dedicated to the extraction of illumination-invariant features from low-light images, complemented by a newly designed loss function that reinforces feature consistency across varying illumination levels. Both modules can be easily integrated into existing object detection frameworks. Our empirical findings reveal a significant improvement in low-light object detection tasks, as well as promising results in both well-lit and over-lit scenarios.
% , and our empirical results demonstrate a notable enhancement in performance for low-light object detection tasks. 
% \textbf{Moreover, we broaden the application of the YOLA to various challenging illumination conditions (over-exposure and normal exposure), further showcasing its superior generalization capabilities.} 
\end{abstract}   
\section{Introduction}
\label{sec:intro}
%In recent years, object detection has made significant advancements in its ability to identify and locate specific objects in digital images or videos~\cite{girshick2014rich,girshick2015fast,ren2015faster,zhu2017flow}. It finds extensive applications across various domains, such as autonomous driving, surveillance systems, face recognition, object tracking, and more~\cite{Li_2023_ICCV,hong2021sspnet,zhang2016joint}. %Despite the significant advancements made in general object detection, low-illumination object detection poses a persistent challenge. Inadequate lighting conditions can lead to poorly illuminated objects, resulting in reduced visibility and potential misdetections~\cite{yang2020advancing,loh2019getting}. 

In the field of computer vision, object detection stands as a cornerstone, driving advancements in numerous applications ranging from autonomous vehicles to security surveillance~\cite{Li_2023_ICCV,zhang2016joint,hong2021sspnet}. The ability to accurately identify and locate objects in digital imagery has seen remarkable progress, largely due to the advent of deep learning techniques~\cite{girshick2014rich,girshick2015fast,ren2015faster}. However, despite these advancements, object detection in low-light conditions remains a significant challenge. Low-light environments lead to poor image quality, reduced visibility, and increased misdetections in night-time surveillance and twilight driving~\cite{yang2020advancing,loh2019getting}.

Traditional methods in tackling low-light object detection have predominantly leaned towards image enhancement techniques~\cite{guo2020zero, jiang2021enlightengan,zhang2019kindling,lv2018mbllen}. While these methods have demonstrated effectiveness in improving visual aesthetics and perceptual quality, they often do not directly translate to improved object detection performance. This discrepancy arises because these enhancement techniques are typically optimized for human visual perception, which does not always correlate with the requirements for effective and accurate object detection by machine learning models.

%These techniques are designed to artificially brighten images, reduce noise, and improve contrast, essentially attempting to convert low-light images into ones that resemble their well-lit counterparts. 

%To address this problem, methods~\cite{guo2020zero, jiang2021enlightengan,zhang2019kindling,lv2018mbllen} based on human vision attempt to restore low-light images to normal lighting conditions. However, this is unnecessary for object detection tasks as these methods optimize based on human visual preferences, which do not align with the objective detection loss.

% Moreover, another line of research investigates finetuning the pre-trained models in the target domain. Detectors are trained on a large volume of normal lit images ~\cite{XXXX} and finetuned on small-size low-light datasets~\cite{XXXX}. To better utilize cross-domain information, MAET ~\cite{XXXX} has designed a framework that enables the network to simultaneously learn common features from both normal and low illumination images. It implicitly models illumination-invariant features through the network. More recently, ~\cite{XXX} have attempted to better extract features from low-illumination images. They employ Laplacian pyramids ~\cite{XXXX} to extract multi-scale edge information and enhance the input images. Additionally, ~\cite{XXX} try to improve the effectiveness by inversing the corrupted images to a normal appearance while training the detector. These methods predefined a set of fixed, human visual-based features that are widely utilized in computer vision and are independent of illumination.

In addition to image enhancement strategies, another research direction involves fine-tuning pre-trained models for low-light conditions. Typically, detectors are initially trained on extensive datasets of well-lit images, such as those from Pascal VOC~\cite{everingham2010pascal} and Microsoft COCO~\cite{lin2014microsoft}, and subsequently fine-tuned on smaller, low-light datasets~\cite{yang2020advancing,loh2019getting}. To enhance the utilization of cross-domain information, the MAET framework~\cite{cui2021multitask} was developed to learn intrinsic visual structure features by separating object features from those caused by degradation in image quality. Similarly, methods~\cite{liu2022image, kalwar2023gdip} aim to restore the normal appearances of corrupted images during detector training. However, these techniques often depend heavily on synthetic datasets, which could limit their real-world applicability.

Recent methods in low-light object detection, such as those in~\cite{qin2022denet, yin2023pe}, use Laplacian pyramids~\cite{burt1987laplacian} for multi-scale edge extraction and image enhancement. FeatEnHancer~\cite{hashmi2023featenhancer} further leverages hierarchical features for improved low-light vision. However, these task-specific, loss-driven approaches often grapple with a larger solution space due to varying illumination effects. 

In this study, we introduce a novel approach that explicitly leverages illumination-invariant features, utilizing the principles of the Lambertian image formation model ~\cite{shafer1985using}. Under the Lambertian assumption, we can express the pixel values in each channel as a discrete combination of three key components: the surface normal, the light direction (both of which are solely related to the pixel's position), the spectral power distribution, and the intrinsic properties of the pixel itself. The illumination-invariant feature can be learned by eliminating the position-related term and spectral power-related term~\cite{gevers1999color}. We introduce this concept of extracting illumination-invariant features into low-light detection tasks and demonstrate that incorporating this feature yields significant performance improvements in low-light detection tasks. We further improve this illumination-invariant feature using task-driven kernels. Our key observation is that by imposing a zero-mean constraint on these kernels, the feature can simultaneously discover richer downstream task-specific patterns and maintain illumination invariance, thereby improving performance.
% We further improve this illumination-invariant feature with learnable convolutional kernels. Our key observation is that by imposing a zero-mean constraint on these kernels, it becomes feasible to dynamically adjust their parameters. This remarkable flexibility enables the learning of a diverse set of kernels, each meticulously capturing unique patterns and textures, thereby greatly facilitating the \textbf{downstream task-specific extraction of features} that remain invariant to changes in illumination.

%, as proposed in PIE-Net~\cite{das2022pie} for intrinsic decomposition tasks

\begin{figure}[t]
   %\captionsetup{type=figure}
   \includegraphics[width=1.0\textwidth]{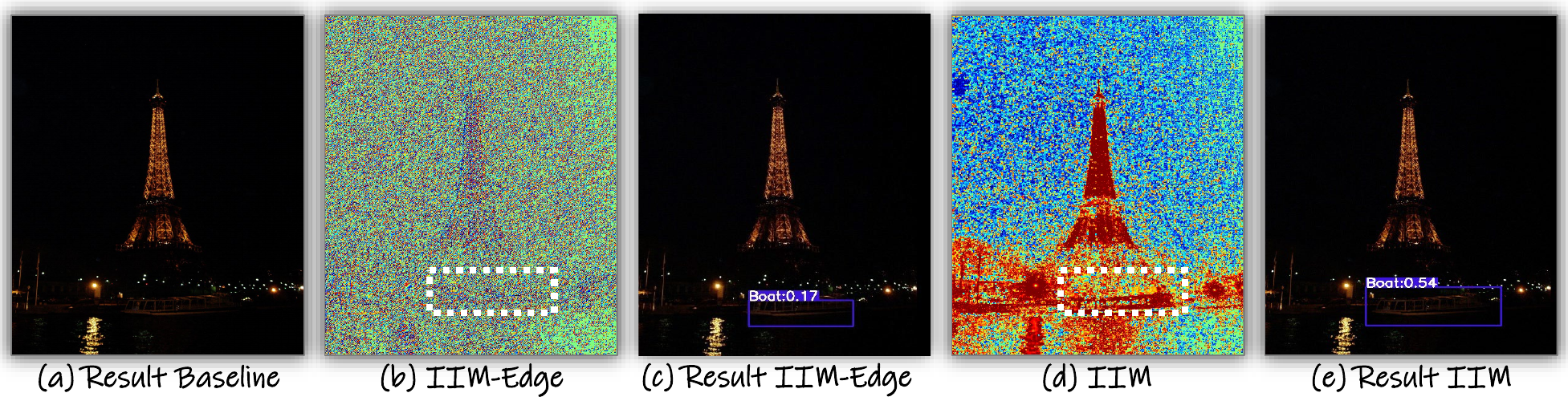}
   \caption{(a): The base detector failed to recognize objects. (b, c) However, when IIM is employed with a simple edge feature, the object is identified. (d, e) Furthermore, the full IIM utilizes a task-driven learnable kernel to extract illumination-invariant features that are richer and more suitable for the detection task than simple edge features.
}
\label{fig:teaser}
\end{figure}

Towards this end, we propose the Illumination-Invariant Module (IIM), a versatile and adaptive component designed to integrate the information gleaned from these specialized kernels with standard RGB images. The IIM can be seamlessly integrated with a variety of existing object detection frameworks, enhancing their capability to perform accurately in low-light environments, whether through naive edge features or diverse illumination-invariant characteristics, as shown in Fig~\ref{fig:teaser}. 
We further conduct experiments on the ExDark and $UG^{2}+$DARK FACE datasets to evaluate our method. Our experimental results demonstrate that the integration of the IIM significantly enhances the detection accuracy of existing methods, leading to substantial improvements in low-light object detection.
To summarize, our contributions are as follows:

\begin{itemize}

\item We introduce YOLA, a novel framework for object detection in low-light conditions by leveraging illumination-invariant features. 
%We introduce YOLA, a novel framework for object detection in low-light scenarios.
\item We design a novel Illumination-Invariant Module to extract illumination-invariant features without requiring additional paired datasets, and can be seamlessly integrated into existing object detection methods. 
\item We provide an in-depth analysis of the extracted illumination-invariant paradigm and propose a learning illumination-invariant paradigm.
% \item To the best of our knowledge, this is the first work to learn illumination-invariant features for low-light object detection. We provide an in-depth analysis of the extracted illumination-invariant paradigm and propose a learning illumination-invariant paradigm.
\item Our experiments show YOLA can significantly improve the detection accuracy of existing methods when dealing with low-light images. 
%Our method outperforms state-of-the-art competitors on both low-light object detection datasets.

\end{itemize}
\begin{figure*}[t]
  \centering
  \vspace{3mm}
  \includegraphics[width=1.0\linewidth]{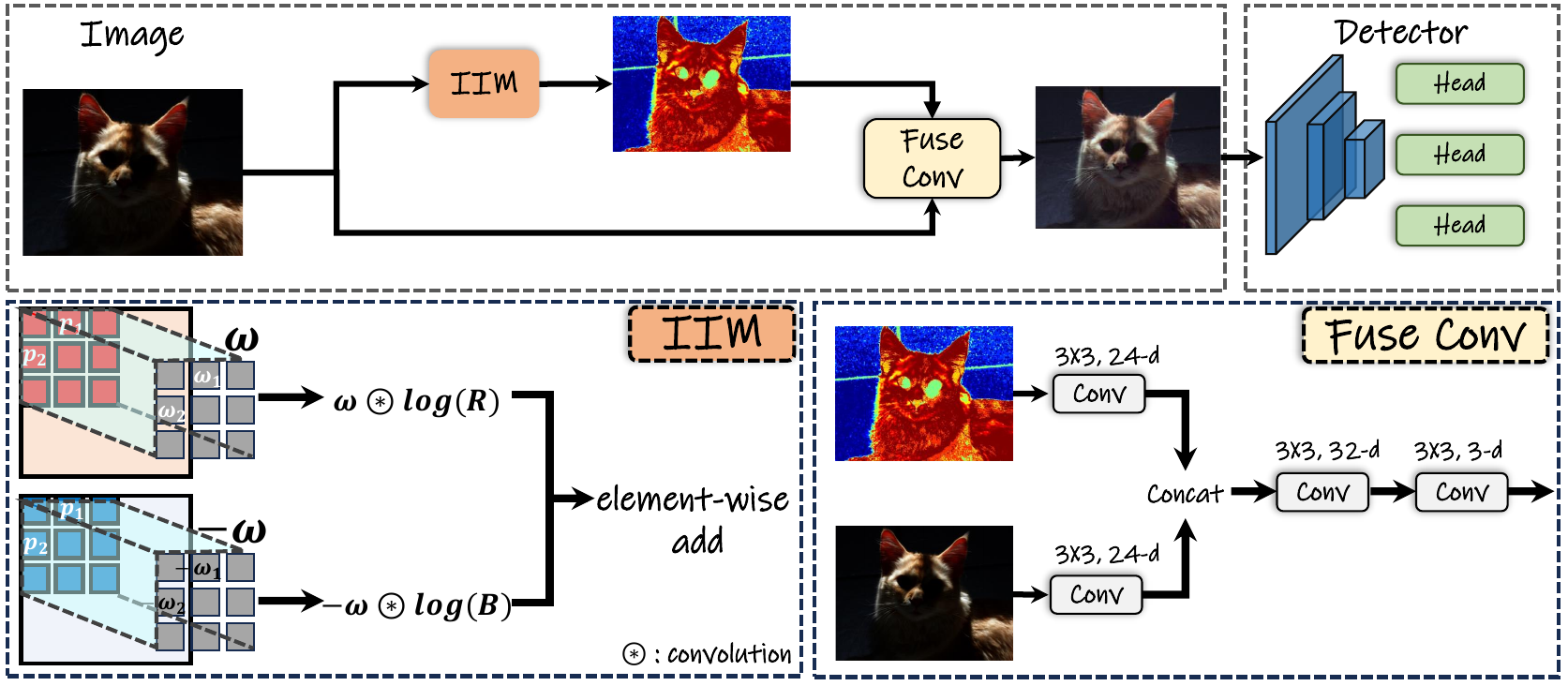}
  %\fbox{\rule{0pt}{2in} \rule{0.9\linewidth}{0pt}}
   %\includegraphics[width=0.8\linewidth]{egfigure.eps}

   \caption{The overall pipeline of YOLA.YOLA extracts illumination-invariant features via IIM and integrates them with original images by leveraging a fuse convolution block for the subsequent detector.}
   \label{fig:pipeline}
   \vspace{-5mm}
\end{figure*}
%-------------------------------------------------------------------------

\section{Related work}
\label{sec:related}

\subsection{General object detection}
Current modern object detection methods can be classified as anchor-based and anchor-free. 
The anchor-based detectors are derived from the sliding-window paradigm, where the dense anchor can be viewed as the sliding-window arranged in spatial space. Subsequently,  the anchors are assigned as positive or negative samples based on the matching strategy (i.e., Intersection-over Union (IoU)~\cite{girshick2014rich}, Top-K~\cite{zhang2020bridging, zhang2020dynamic}).
Common anchor-based methods include R-CNN~\cite{girshick2014rich,girshick2015fast,ren2015faster}, SSD~\cite{liu2016ssd}, YOLOv2~\cite{redmon2017yolo9000}, and RetinaNet~\cite{lin2017focal}, among others. In contrast, the anchor-free detectors liberate the handcraft anchor hyper-parameter setting, enhancing their potential in terms of generalization capability. Prominent methods in anchor-free include YOLOv1~\cite{redmon2016you}, FCOS~\cite{tian2019fcos},  and DETR~\cite{carion2020end}.
Despite the remarkable achievements of both anchor-based and anchor-free detectors in general object detection, they exhibit unsatisfactory performance under low-light conditions.
\vspace{-5mm}
\subsection{Low-light object detection}
% Object detection in low-light conditions remains a significant challenge. One common line of research involves leveraging image enhancement techniques, such as MBLLEN~\cite{guo2020zero}, KIND~\cite{zhang2019kindling}, RUAS~\cite{liu2021retinex}, and others~\cite{guo2020zero, jiang2021enlightengan,jiang2023low}  to directly improve the quality of the low-light image.  
Object detection in low-light conditions remains a significant challenge. One common line of research involves leveraging image enhancement techniques, such as KIND~\cite{zhang2019kindling}, SMG~\cite{xu2023low}, NeRCo~\cite{ yang2023implicit}, and others~\cite{guo2020zero, jiang2021enlightengan,jiang2023low,jiang2024lightendiffusion}  to directly improve the quality of the low-light image.
The enhanced images are then deployed in the subsequent training and testing stages of detection.  However, the objective of image enhancement is inherently different from that of object detection, making this strategy suboptimal. To address this, some researchers~\cite{huang2020dsnet, cui2022you} explore integrating image enhancement with object detection during the training process. Nevertheless, the task of balancing hyperparameters to equilibrate visual quality and detection performance remains intricate. Recently, Sun et al.~\cite{sun2022rethinking} proposed a targeted adversarial attack paradigm aimed at restoring degraded images to ones that are more favorable for object detection.
MAET~\cite{cui2021multitask} trained on a low-light synthetic dataset, obtaining the pre-trained model endowed intrinsic structure decomposition ability for downstream lowlight object detection.
% MAET~\cite{cui2021multitask} performs supervised multi-task learning by constructing a synthetic dataset, obtaining the pre-trained model toward intrinsic representation for downstream lowlight object detection.
Further, IA-YOLO~\cite{liu2022image} and GDIP~\cite{kalwar2023gdip} elaborately design the differentiable image processing module to enhance image adaptively for adverse weather object detection.
% Note that the aforementioned methods either require a dedicated low-light enhancement dataset or rely heavily on synthetic datasets in training, which potentially limits the model's generalization ability in real-world scenarios. 
Note that the aforementioned methods either require a dedicated low-light enhancement dataset or rely heavily on synthetic datasets in training.
To mitigate the limitations,  a set of methodologies~\cite{qin2022denet,yin2023pe,hashmi2023featenhancer} utilize multi-scale hierarchical features and are driven purely by task-specific loss to improve low-light vision.
Unlike those methods, we introduce illumination-invariant features to alleviate the effect of illumination on low-light object detection, without requiring additional low-light enhancement datasets or synthetic datasets.

\subsection{Illumination invariant representation}
Adverse illumination typically degrades the performance on downstream tasks, prompting researchers to explore illumination-invariant techniques to mitigate this impact.
For high-level tasks, Wang et al.~\cite{wang2011illumination} proposed an illumination normalization method for Face Recognition.
Alshammari et al.~\cite{alshammari2018impact} use illumination-invariant image representation to improve automotive scene understanding and segmentation. Lu et al.~\cite{lu2023cross} convert RGB images to illumination-invariant chromaticity space, preparing for the following feature extraction to achieve traffic object detection in various illumination conditions.
For low-level tasks, several physics-based invariants, such as Colour Ratios~\cite{finlayson1992colour} (CR) and Cross Colour Ratios~\cite{gevers1999color} (CCR), are employed to decompose the illumination for intrinsic image decomposition~\cite{das2022pie, das2022intrinsic,das2023idtransformer}.
However, these methods leverage illumination-invariant representations derived from the fixed formulations, which may not adequately capture the diverse and complex illumination scenarios that are specific to downstream applications. In contrast,  our method enables the adaptive learning of illumination-invariant features in an end-to-end manner, thereby enhancing compatibility with downstream tasks.
% However, these methods leverage illumination-invariant representations derived from the fixed formulations, which may not adequately capture the diverse and complex illumination scenarios. In contrast,  our method enables the illumination-invariant features can be adaptively learned in an end-to-end fashion.

\section{Method}
\label{sec:Method}
In this section, we formally introduce YOLA, a novel method for low-light object detection. As illustrated in Fig.\ref{fig:pipeline}, the key component of YOLA is the Illumination Invariant Module (IIM) focusing on feature learning to derive downstream task-specific illumination-invariant features. These features can be integrated with existing detection modules, enhancing their capability in low-light conditions.  Next, we will introduce the derivation of illumination-invariant features and provide a detailed description of IIM's specific implementation.

\subsection{Illumination invariant feature}
\label{sec:IIF}
\paragraph{\textbf{Notation}:}  Let $I$ represents an image in the standard RGB domain, and let $C \in [R, G, B]$ represent the image in the red, green, or blue channel. We define the value in channel $C$ of a pixel $p_i$ as $C_{p_i}$, where $i \in I$ is the pixel index.

%Let $I = [R, G, B]$ represent an image in the standard RGB domain, and let $C^{\in H \times W} \in [R, G, B]$ represent the image in the red, green, or blue channel. Here, $H$ and $W$ represent the height and width of the image, respectively. We can define the value of $C$ at point $p_i$ as $C_{p_i}$, where $i \leq HW$.

\paragraph{\textbf{Lambertian assumption}:} According to body reflection term of the dichromatic reflection model, the value of $C_{p_{i}}$ can be expressed in the discreet form as follows:
% According to previous work PIE-Net ~\cite{das2022pie}, the value of $C_{p_{i}}$ can be expressed in the discreet form as follows:
    \begin{equation}
        C_{p_{i}} = m(\Vec{n}_{p_{i}}, \Vec{l}_{p_{i}})e^{C_{p_{i}}}(\lambda)\rho^{C_{p_{i}}}(\lambda),
    \label{equ:image_dct}
    \end{equation}
Here, $\Vec{n}_{p_{i}}, \Vec{l}_{p_{i}}$ represents  surface normal and light direction respectively, and $m$ denotes the interaction function between them. The term $e^{C_{p_{i}}}$ represents the spectral power distribution of the illuminant at point $p_{i}$ in color channel $C$
, and $\rho^{C_{p_{i}}}$ represents the intrinsic property (reflectance) of the object at point $p_{i}$ in color channel $C$. 

It becomes apparent that the term $m$ is determined solely by the positional component, with no impact from the color channels. This observation leads to the strategy of calculating the difference between values of different color channels at the same spatial positions to effectively eliminate the influence of $m$. To eliminate the term $e$, we can utilize the assumption that illumination is approximately uniform across adjacent pixels. Consequently, by computing the difference between values of neighboring pixels, we can further further eliminate the influence of $m$. 
% Towards this end, PIE-Net ~\cite{das2022pie} proposed a Cross Color Ratio based procedure for Illumination Invariant gradient calculation.

%We can observe that the term $m$ is exclusively influenced by the positional component and remains unaffected by color channels. This implies that by computing the difference between values from distinct color channels at the same spatial location, one can effectively eliminate the influence of $m$. As for term $e$, we can leverage the assumption that the illuminates between adjacent pixels is approximately equal and by computing the difference between values of nearby pixels, one can further eliminate the influence of $m$.

\paragraph{\textbf{Cross color ratio}:} Taking into consideration two adjacent pixels, denoted as $p_1$ and $p_2$, along with the red ($R$) and blue ($B$) channels, we can determine the ratio $M_{rb}$ between the red and blue channels through the following computational procedure:
\begin{equation}
    M_{rb} = \frac{R_{p_{1}}B_{p_{2}}}{R_{p_{2}}B_{p_{1}}}.
\label{equ:ccr}
\end{equation}
Taking the logarithm of $M_{rb}$ and substituting the pixel values with Eq.~\ref{equ:image_dct}, we get:
%\begin{equation}
%\label{equ:log_ccr}
%\begin{split}
%    log(M_{rb}) = log(R_{p_{1}}) - log(R_{p_{2}})\\
%    + log(B_{p_{2}})
%      - log(B_{p_{1}}).
%\end{split}
%\end{equation}
%By substituting the pixel values in Eq.~\ref{equ:log_ccr} with Eq.~\ref{equ:image_dct}, we obtain:
\begin{equation}
\begin{split}
    log(M_{rb}) = log(m(\Vec{n_{p_{1}}}, \Vec{l_{p_{1}}}))
     - log(m(\Vec{n_{p_{1}}}, \Vec{l_{p_{1}}}))\\
    + log(e^{R_{p_{1}}}(\lambda))- log(e^{R_{p_{2}}}(\lambda)) \\
    + log(\rho^{R_{p_{1}}}(\lambda))-log(\rho^{R_{p_{2}}}(\lambda))\\
    + log(m(\Vec{n_{p_{2}}}, \Vec{l_{p_{2}}})) -
    log(m(\Vec{n_{p_{2}}}, \Vec{l_{p_{2}}})) \\
    + log(e^{B_{p_{2}}}(\lambda)) -log(e^{B_{p_{1}}}(\lambda)) \\
    +log(\rho^{B_{p_{2}}}(\lambda))-log(\rho^{B_{p_{1}}}(\lambda)).
\end{split}
\label{equ:ccr_detail}
\end{equation}

With the illumination assumption that $e^{C_{p_{1}}} \approx	 e^{C_{p_{2}}}$,  the above equation can be further simplified into an illumination-invariant form:

%In the above equation, we are able to eliminate the term $m$. As for the remaining term $e$, we can leverage the approximation that the illuminance between adjacent pixels is approximately equal:
%\begin{equation}
%e^{C_{p_{1}}} = e^{C_{p_{2}}}.    
%\end{equation}
%Consequently, the Eq.~\ref{equ:log_ccr} can be further simplified as follows:
\begin{equation}
\begin{split}
    log(M_{rb}) = 
    log(\rho^{R_{p_{1}}}(\lambda))- log(\rho^{R_{p_{2}}}(\lambda))\\
    +log(\rho^{B_{p_{2}}}(\lambda))-log(\rho^{B_{p_{1}}}(\lambda))
\end{split}
\label{equ:ccr_simple}
\end{equation}

By observing the elimination in Eq.~\ref{equ:ccr_simple}, we can find that subtraction within the \textbf{same channel} eliminates the illumination term (implemented by zero-mean constraint), while \textbf{cross-channel} subtraction removes the surface normal and light direction terms,  which motivates us to design the learning illumination-invariant paradigms.

In this case, we can use a convolution operation to extract features, as shown in Fig.~\ref{fig:pipeline}. The extracted features are processed and fused by the IIM before being sent to the detector. When using fixed weights of adjacent pixels with a subtraction value of 1 or $-1$, we refer to it as IIM-Edge. Next, we will provide a detailed introduction to the IIM.

\subsection{Illumination invariant module}
While Eq. \ref{equ:ccr_simple} offers a straightforward and effective method for calculating Illumination Invariant features, its rigidity presents certain limitations. 
Specifically, the fixed nature of this equation may not adequately capture the diverse and complex variations in illumination that are specific to downstream tasks across different scenarios.
To address this, we have evolved the equation into a more adaptable form using convolutional operations. Instead of relying on a single kernel, our approach involves learning a set of convolutional kernels. This strategy not only enhances the robustness of the Illumination Invariant feature extraction but also improves its efficiency. 
To this end, we propose Illumination Invariant Module comprising two main components, including learnable kernels and a zero-mean constraint.
Note that Illumination Invariant Module yield features are inherently illumination invariant at initialization. Subsequent kernel learning is geared towards producing task-specific illumination invariant features for downstream tasks.

%The key innovations in our IIM are these learnable kernels, coupled with an additional zero-mean constraint, which collectively form the backbone of our enhanced feature extraction process.

\paragraph{\textbf{Learnable kernel}.} The goal is to transform the fixed illumination-invariant feature into a learnable form. Specifically,  we aim to learn a set of convolutional kernels ${\mathcal{W}_{1}, \mathcal{W}_{2}, \cdots \mathcal{W}_{n}^{\in k \times k}} $, where $n$ represents the number of kernels and $k$ denotes the kernel size. Here, we extend the fixed feature into a more versatile and generalized form. Let $p_{i}$ and $w_{i}$ represent a group pixel position and its corresponding weight within a kernel $\mathcal{W}_{n}$, where $i=0,1,\cdots k^{2}$. These parameters enable us to evolve the Cross Color Ratio (CCR) into an adaptable form, enhancing its capability to effectively handle varying illumination conditions. Note that $w_{i}$ is trainable, rendering the positive or negative polarity inconsequential.

\begin{equation}
    M_{rb} = \prod \limits_{i=1}^{k^2-1} \left(\frac{R_{p_{i}}}{B_{p_{i}}}\right)^{w_i}\left(\frac{B_{p_{i+1}}}{R_{p_{i+1}}}\right)^{w_{i+1}} = \prod \limits_{i=1}^{k^2} \left(\frac{R_{p_{i}}}{B_{p_{i}}}\right)^{w_i}
    \label{equ:lccr}
\end{equation}

To make the extended form still satisfy Illumination Invariant, the logarithm of $M_{rb}$ should satisfy the following constraints:
\begin{equation}
\begin{split}
    \begin{cases}
    \sum_{i}^{k^2} w_{i}log(e^{R_{p_{i}}}(\lambda)) = 0 \\
    \sum_{i}^{k^2} w_{i}log(e^{B_{p_{i}}}(\lambda)) = 0
    \end{cases}
\end{split}
\label{equ:lccr_simple_cons}
\end{equation}
If the above equation holds true, the $e$ term and the $m$ term are eliminated. The final feature can be expressed in a generalized form:
\begin{equation}
\begin{split}
    log(M_{rb}) = \sum_{i}^{k^2}
    w_{i}log(\rho^{R_{p_{i}}}(\lambda)) -\sum_{i}^{k^2}w_{i}log(\rho^{B_{p_{i}}}(\lambda))
\end{split}
\label{equ:lccr_simple_conss}
\end{equation}
Similarly, we can obtain $log(M_{rg})$ and $log(M_{gb})$ to form $f_{\mathcal{W}_{i}}( I)$.

The resulting features obtained by applying the kernel $\mathcal{W}_{i}$ to the image $I$ denoted as $f_{\mathcal{W}_{i}}( I)$, can be expressed as:

% % \left[\begin{array}{l}
% % F_{rb}^{i} \\
% % F_{rg}^{i} \\
% % F_{gb}^{i}
% % \end{array}\right]=

\begin{equation}
\mathclap{
f_{\mathcal{W}_{i}}( I) =\left[\begin{array}{l}
\mathcal{W}_{i} \circledast log(R) + (-\mathcal{W}_{i}) \circledast log(B) \\
\mathcal{W}_{i} \circledast log(R) + (-\mathcal{W}_{i}) \circledast log(G) \\
\mathcal{W}_{i} \circledast log(G) + (-\mathcal{W}_{i}) \circledast log(B)
\end{array}\right]
}
\end{equation}
where $\circledast$ denotes the convolution. 

% However, as delineated in Eq \ref{equ:ccr_detail}, the primary challenge we face is the elimination of terms related to illumination within this learnable framework. To address this, we introduce two learnable parameters $w_1$ and $w_2$. 

\paragraph{\textbf{Zero mean constraint}:} 
% Based on Eq.~\ref{equ:lccr_simple_cons}, we can set $w_1$ equal to $w_2$ to satisfy the equation. Generalizing this to the case of convolutional kernels, we enforce the mean of $\mathcal{W}_{n}^{\in k \times k}$ to be $0$ by employing the following constrain:
Drawing from Eq.~\ref{equ:lccr_simple_cons} and the approximation $e^{R_{p_{1i}}} \approx	 e^{B_{p_{i}}}$, in the context of convolutional kernels, we simply ensure that the mean of $\mathcal{W}_{n}^{\in k \times k}$ to be $0$, as depicted by:
% \begin{equation}
%  \mathcal{W}_{n} - \overline{\mathcal{W}},
% \end{equation}
% \begin{figure*}[t]
% \vspace{5mm}
%   \centering
%   \includegraphics[width=1.\linewidth, height=0.7\linewidth]{figs/eccv_visualization.pdf}
%    \caption{Qualitative comparisons of TOOD detector on both ExDark and $UG^{2}+$DARK FACE dataset, where the top 2 rows visualize the detection results from ExDark, and the bottom 2 rows show the results from $UG^{2}+$DARK FACE. The images are being replaced with enhanced images generated by LLIE or low-light object methods. Red dash boxes highlight the inconspicuous cases. Zoom in red dash boxes for the best view.}
%    \label{fig:qualitative}
%    \vspace{-2mm}
% \end{figure*}
\begin{equation}
 %\overline{\mathcal{W}_{n}} = 0
 \overline{\mathcal{W}_{n}} = \frac{1}{k^2} \sum_{i=1}^{k^2} w_{i} = 0
\end{equation}
This constraint is enforced by substituting the mean value from the kernel $\mathcal{W}_{n} = \mathcal{W}_{n} - \overline{\mathcal{W}_{n}}$.  %here $\overline{\mathcal{W}_{n}}$ is the mean value of $\mathcal{W}_{n}$

\begin{figure*}[t]
\vspace{5mm}
  \centering
  \includegraphics[width=1.\linewidth, height=0.7\linewidth]{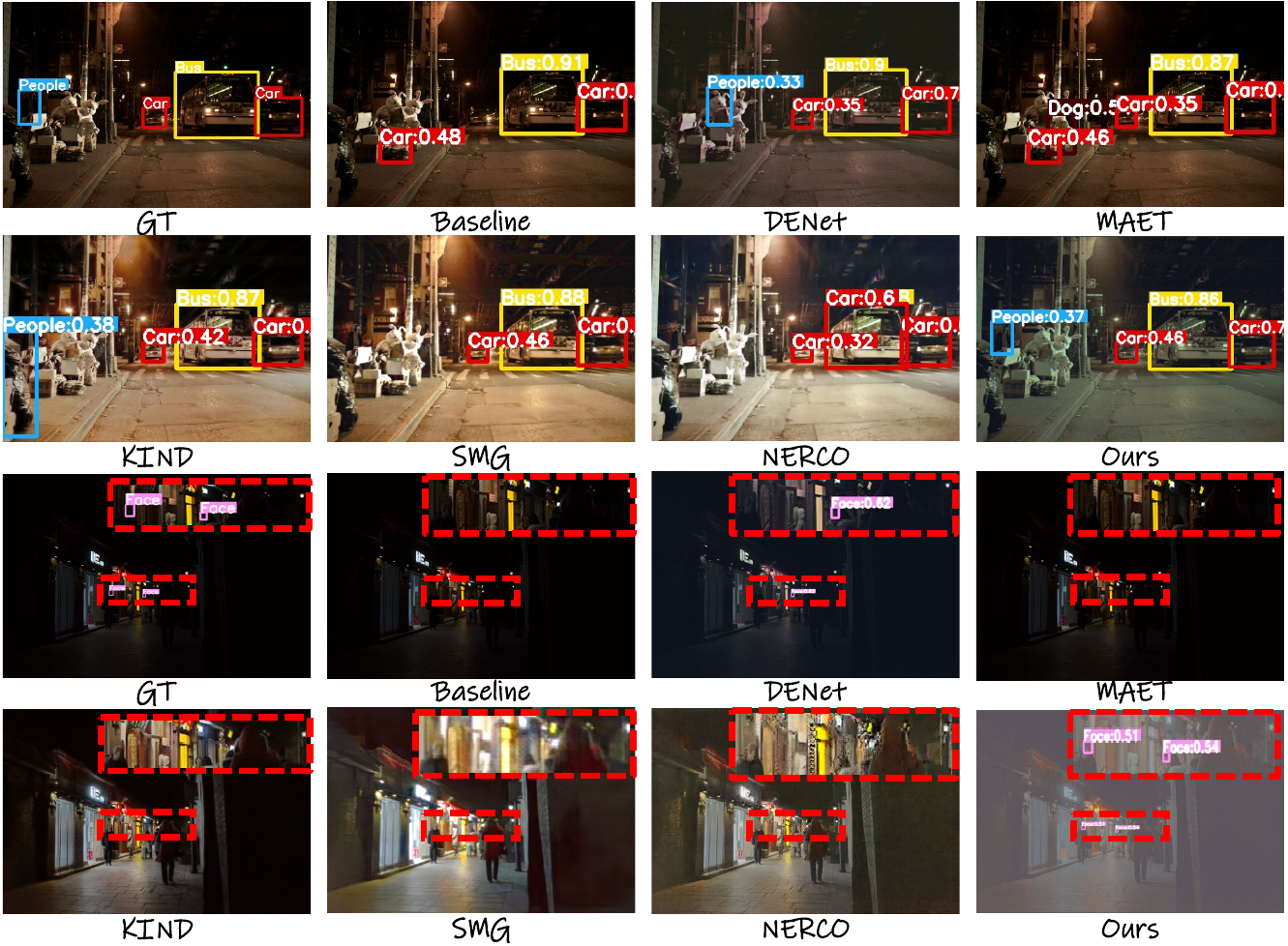}
   \caption{Qualitative comparisons of TOOD detector on both ExDark and $UG^{2}+$DARK FACE dataset, where the top 2 rows visualize the detection results from ExDark, and the bottom 2 rows show the results from $UG^{2}+$DARK FACE. The images are being replaced with enhanced images generated by LLIE or low-light object methods. Red dash boxes highlight the inconspicuous cases. Zoom in red dash boxes for the best view.}
   \label{fig:qualitative}
   \vspace{-2mm}
\end{figure*}
\section{Experiments}

\subsection{Implementation details}
We evaluate the proposed method using the popular anchor-based detector YOLOv3~\cite{redmon2018yolov3} and the anchor-free detector TOOD~\cite{feng2021tood}.
Both detectors are initially pre-trained on the COCO dataset and subsequently fine-tuned on the target datasets utilizing the SGD~\cite{ruder2016overview} optimizer with an initial learning rate of $1e$-$3$.
Specifically, we resize the ExDark dataset images to $608 \times 608$ and train both detectors for 24 epochs, reducing the learning rate by a factor of 10 at epochs 18 and 23. 
For the $UG^{2}+$DARK FACE dataset, we resize images to $1500 \times 1000$ for TOOD and maintain the $608 \times 608$ resolution for YOLOv3 to be consistent with MAET.
YOLOv3 is trained for $20$ epochs, with the learning rate decreased by a factor of $10$ at $14$ and $18$ epochs. TOOD are trained for $12$ epochs, with the learning rate decreased by a factor of $10$ at $8$ and $11$ epochs.
Additionally, we implement a straightforward illumination-invariant model, denoted as \textbf{YOLA-Naive}, by removing the IIM and ensuring various illumination features are consistently imposed by an MSE loss.
We implement YOLA using the MMDetection toolbox~\cite{chen2019mmdetection}.
\begin{table*}
\vspace{-5mm}
\begin{minipage}[t]{0.48\linewidth}
\centering
\footnotesize
\begin{adjustbox}{width=\linewidth}
 \begin{tabular}{c c>{\columncolor{gray!20}} cc >{\columncolor{gray!20}} c}
    \multirow{2}{*}{Methods} &
      \multicolumn{2}{c}{YOLOv3} &
      \multicolumn{2}{c}{TOOD}\\
      % \vspace{2pt}
      \cmidrule(lr){2-3}\cmidrule(lr){4-5}
      & recall & {mAP$_{50}$} & recall & {mAP$_{50}$}\\
   \toprule
    Baseline                      & 84.6 & 71.0 & 91.9  & 72.5\\
    KIND~\cite{zhang2019kindling} & 83.3 & 69.4 & 92.1  & 72.6 \\
    % MBLLEN~\cite{lv2018mbllen}    & 83.2 & 70.3 & 92.4  & 73.5 \\
    % RAUS~\cite{liu2021retinex}    & 81.6 & 65.7 & 91.0  & 69.4  \\
    SMG~\cite{xu2023low}    & 82.3 & 68.5 & 91.8  & 71.5  \\
    NeRCo~\cite{yang2023implicit}  & 83.4 & 68.5 & 91.8  & 71.8  \\
    DENet~\cite{qin2022denet}     & 84.2 & 71.3 & 92.6  & 73.5\\
    GDIP~\cite{zhang2019kindling} & 84.8 & 72.4 & 92.2  & 72.8 \\
    IAT~\cite{zhang2019kindling}   & 85.0 & 72.6 & 92.9  & 73.0 \\
    MAET~\cite{cui2021multitask}  & 85.1 & 72.5 & 92.5  & 74.3 \\ \hline
    YOLA-Naive  & 84.8 & 71.6 & 91.8  & 71.6 \\ 
    \textbf{YOLA}                 & \textbf{86.1} &\textbf{72.7} & \textbf{93.8}  & \textbf{75.2}\\
    \bottomrule
  \end{tabular}
  \end{adjustbox}
 \caption{Quantitative comparisons of the ExDark dataset based on YOLOv3 and TOOD detectors.}
 \label{tab:results_exdark}
\end{minipage}
\hfill
\begin{minipage}[t]{0.48\linewidth}
\centering
\footnotesize
\begin{adjustbox}{width=\linewidth}
 \begin{tabular}{c c>{\columncolor{gray!20}} cc >{\columncolor{gray!20}} c}
    \multirow{2}{*}{Methods} &
      \multicolumn{2}{c}{YOLOv3} &
      \multicolumn{2}{c}{TOOD}\\
      % \vspace{2pt}
      \cmidrule(lr){2-3}\cmidrule(lr){4-5}
     & recall & {mAP$_{50}$} & recall & {mAP$_{50}$}\\
   \toprule
    Baseline                      & 77.9 & 60.0 & 81.5  & 62.1\\
    KIND~\cite{zhang2019kindling} & 76.0 & 58.4 & 82.4  & 63.8 \\
    % MBLLEN~\cite{lv2018mbllen}    & 76.1 & 58.2 & 82.6  & 65.3 \\
    % RAUS~\cite{liu2021retinex}    & 76.9 & 59.3 & 82.7  & 64.8  \\
    SMG~\cite{xu2023low}   & 69.3 & 48.9 & 77.1  & 55.8  \\
    NeRCo~\cite{yang2023implicit} & 68.9 & 49.1 & 76.8  & 55.6  \\
    DENet~\cite{qin2022denet}     & 77.7 & 60.0 & 84.1  & 66.2\\
    GDIP~\cite{zhang2019kindling}  & 77.8 & 60.4 & 82.1  & 62.9 \\
    IAT~\cite{zhang2019kindling}   & 77.6 & 59.8 & 82.1  & 62.0 \\
    MAET~\cite{cui2021multitask}  & 77.9 & 59.9 & 83.6  & 64.8 \\ \hline
    YOLA-Naive  & 76.6 & 59.2 & 82.8  & 64.6 \\ 
    \textbf{YOLA}                 & \textbf{79.1} &\textbf{61.5} & \textbf{84.9}  & \textbf{67.4}\\

    \bottomrule
  \end{tabular}
\end{adjustbox}
\caption{Quantitative comparisons of the $UG^{2}+$DARK FACE dataset based on YOLOv3 and TOOD detectors.}
\label{tab:results_dark_face}
\end{minipage}
\vspace{-8mm}
\end{table*}

\subsection{Dataset}
We evaluate our proposed method on both real-world scenarios datasets: exclusively dark~\cite{loh2019getting} (ExDark) and $UG^{2}+$DARK FACE~\cite{yang2020advancing}. ExDark dataset contains 7363 images ranging from low-light environments to twilight, including 12 categories, 3,000 images for training, 1,800 images for validation, and 2,563 images for testing.
We calculate the overall mean average precision (mAP$_{50}$) and mean recall at the IoU threshold of 0.5 as the evaluation metric.  $UG^{2}+$DARK FACE dataset contains 6,000 labeled face bounding box images, where 5,400 images are allocated for training and  600 images are reserved for testing, and calculating the corresponding recall and  mAP$_{50}$ as evaluation metrics. Additionally, we also evaluate the generalization ability of our method on the COCO 2017~\cite{lin2014microsoft} dataset.
\label{sec:experiment}

\subsection{Low-light object detection}
Table~\ref{tab:results_exdark} presents the quantitative results of YOLOv3 and TOOD detectors on the ExDark dataset, respectively. We report the low-light image enhancement (LLIE) methods, including KIND, SMG, and NeRCo, along with the state-of-the-art low-light object detection methods, DENet, and MAET.
Compared to the low-light object detection methods, the LLIE methods fail to achieve satisfactory performance due to inconsistency between human visual and machine perception. The enhancement methodologies prioritize human preferences. However, it is important to note that optimizing for enhanced visual appeal may not align with optimized object detection performance. Despite being the current state-of-the-art in image enhancement techniques, SMG and NeRCo exhibit worse performance compared to KIND when evaluated in the context of object detection tasks. In contrast, end-to-end approaches such as DENet and MAET, which account for machine perception, generally yield superior results in object detection compared to the LLIE methods. Nevertheless, our method remains simple and effective when compared to similar approaches in the same category. 
Moreover, compared to YOLA-Naive, YOLA exhibits superior performance because its extracted features inherently possess illumination invariance, implying a smaller solution space compared to YOLA-Naive.
Specifically, our method achieves the best performance on both anchor-based YOLOv3 and anchor-free detectors TOOD, surpassing the baseline by significant gains of 1.7 and 2.5~mAP, indicative of its superiority and effectiveness. 
Meanwhile, compared with most LLIE and lowlight object detection techniques, the number of parameters in our YOLA ({$\textbf{0.008M}$}) is significantly lower,  as presented in Table~\ref{tab:params}. This highlights the potential for our method to be deployed in lightweight practical applications.
For a more detailed quantitative comparison, please refer to our appendix.

\subsection{Low-light face detection}
We have shown the results on the ExDark dataset. Next, we showcase the results on a dataset that includes small-sized objects. Table~\ref{tab:results_dark_face} presents the quantitative results of the detector YOLOv3 and TOOD on $UG^{2}+$DARK FACE dataset. 
Significantly, it is worth noting that most LLIE methods integrated into the YOLOv3 detector fail to achieve satisfactory results. This implies that the utilization of enhancement-based approaches can impair the details of small faces, hindering the learning of useful representations in such images. On the other hand, methods considering the object detection task demonstrate better performance, where YOLA increases the 1.5~mAP, demonstrating its superior performance and generalization capability.
For the recently advanced detector TOOD, our method still outperforms these LLIE and low-light object detection methods, achieving a remarkable mAP of 67.4.
This underscores YOLA's superior generalization capabilities in improving the performance of both anchor-based and anchor-free detection paradigms.
\begin{table}[tbp]
\vspace{-5mm}
\begin{minipage}[t]{.45\linewidth}
\begin{adjustbox}{width=1.05\linewidth}
\renewcommand\arraystretch{1.05}
\begin{tabular}{c|cccc}
\toprule
\multirow{1}{*}{Dataset}    & IIM         & {IIM-Edge}             & $\mathcal{Z}_{mean}$  & mAP$_{50}$  \\ 
% & & & \\
% & & & \\
\hline
\multirow{4}{*}{Exdark}     &             &                       &                       & 72.5 \\
                            &             &  \checkmark           &                       & 73.8 \\
                            & \checkmark  &                       &                       & 74.7 \\
                            & \checkmark  &                       &  \checkmark           & \textbf{75.2} \\ \cline{1-5}
\multirow{4}{*}{DarkFace}   &             &                       &                       & 62.1 \\
                            &             & \checkmark            &                       & 64.5 \\
                            & \checkmark  &                       &                       & 66.9 \\    
                            & \checkmark  &                       &   \checkmark          & \textbf{67.4} \\ \hline
\end{tabular}
\end{adjustbox}
\caption{The effectiveness of IIM, IIM-Edge and the zero mean constraint $\mathcal{Z}_{mean}$ based on TOOD. The blank line denotes the baseline.}
\label{tab:tood_fixed}
\end{minipage}
\hspace{0.05\linewidth} % 添加水平间距
\begin{minipage}[t]{0.5\linewidth}
\begin{minipage}[l]{1\linewidth}
\begin{adjustbox}{width=\linewidth}
\begin{tabular}{c|c|lll}
\toprule
Dataset & Method    & AP$_{50}$ & AP$_{75}$ & mAP  \\ \midrule
\multirow{2}{*}{well-lit} & TOOD     & 59.0 & 45.3 & 41.7 \\
                            & + YOLA   & \textbf{59.4} & \textbf{46.0} & \textbf{42.3}  \\ 
\midrule
\multirow{2}{*}{over-light} & TOOD      & 57.4 & 43.8 & 40.5 \\
                            & + YOLA     & \textbf{58.3} & \textbf{44.6} & \textbf{41.2}  \\ 
\bottomrule
\end{tabular}
\end{adjustbox}
\captionsetup{type=table}
\caption{Ablation study for YOLA on COCO {\ttfamily 2017val}.}
\label{tab:coco2017}
\end{minipage}
\begin{minipage}{1\linewidth}
\begin{adjustbox}{width=1.0\linewidth}
\begin{tabular}{l|c|c|c|c|c|l}
\toprule
Method  & Kind & SMG   & NeRco & DENet & MAET & Ours  \\ \hline
Size(M) & 8.21 & 17.90 & 23.30 & 0.04  & 40   & \textbf{0.008}  \\
\bottomrule
\end{tabular}
\end{adjustbox}
\captionsetup{type=table}
 \caption{Model size of different methods.}
  \label{tab:params}
\end{minipage}
\end{minipage}
\vspace{-8mm}
\end{table}

\subsection{Quantitative results}
The top 2 rows of Figure~\ref{fig:qualitative} show the qualitative results from the ExDark dataset using the TOOD detector, where existing methods exhibit missed detections, highlighted by the red dashed boxes.
% The top 2 rows in Fig.~\ref{fig:qualitative} present the qualitative results of the ExDark dataset using the TOOD detector. Existing methods demonstrate missed detections, as indicated by the red dash boxes.
In contrast, YOLA excels in detecting these challenging cases, demonstrating its superior performance in complex scenarios. The bottom 2 rows exhibit the qualitative results of the $UG^{2}+$DARK FACE dataset using the TOOD detector. These faces are typically tiny under low-light conditions, making it difficult for most methods to achieve comprehensive results.

Although our method does not explicitly constrain image brightness, the enhanced images tend to display increased brightness in the final results. The visual results shown in the figures may appear slightly grayish due to the absence of value range constraints on the enhanced images. For image display, we conducted channel-wise normalization.

\subsection{Ablation studies}
\subsubsection{Illumination invariant module}
\label{sec:exp_IIM}
We evaluate the effectiveness of the IIM in detectors TOOD, as presented in Table~\ref{tab:tood_fixed}, respectively. The 1st and 5th rows of Table~\ref{tab:tood_fixed} show the baseline detectors evaluated on ExDark and $UG^{2}+$DARK FACE  dataset. By incorporating the IIM to introduce illumination-invariant features, the detector yields considerable performance gains (2.3 and 4.8~mAP for ExDark and $UG^{2}+$DARK FACE, respectively).

\subsubsection{Zero mean constraint}
\label{sec:exp_ZMC}
By imposing a zero mean constraint on the convolutional kernels, the subtraction formed by the kernels can factor out the illumination items. To evaluate the impact of this constraint, we exclude it from IIM, and the results are shown in Table~\ref{tab:tood_fixed}.
It is evident that the removal of this constraint leads to a decline in performance, with reductions of 0.3 and 0.5~mAP for TOOD. These results indicate that utilizing the zero mean constraint to mitigate the effects of illumination is beneficial to low-light object detection.
% \vspace{-4mm}
\subsubsection{Learnable kernel}
\label{sec:lean_kernel}
The IIM is formed with the learnable kernels, encouraging the illumination-invariant features that are adaptively learned in an end-to-end fashion. 
% In this experiment, we employ the fixed kernels (as specified in Eq.~\ref{equ:ccr_simple}) to form IIM, the results of which are shown in Table~\ref{tab:tood_fixed}. 
In this experiment, we evaluate the fixed kernels (as specified in Eq.~\ref{equ:ccr_simple}, also referred to as IIM-Edge), the results of which are shown in Table~\ref{tab:tood_fixed}. 
It outperforms the baseline by 1.3~mAP on ExDark and 2.4~mAP on $UG^{2}+$DARK FACE, which demonstrates that the incorporation of illumination-invariant features is beneficial for low-light object detection.
Subsequently, we substitute the fixed kernels with the learnable kernels, yielding further gains of 1.4~mAP on ExDark and 2.9~mAP on $UG^{2}+$DARK FACE. These results clearly prove the effectiveness of learnable kernels.
In addition,  we also impose a consistency loss for IIM's output feature to stabilize the kernel learning to prevent trivial solutions within the kernel, mitigating the impact of uneven lighting. (please refer to the appendix~\ref{part:IILoss} for details).

\begin{figure}[t]
\vspace{-5mm}
\centering
   \includegraphics[width=0.92\linewidth, height=0.35\linewidth]{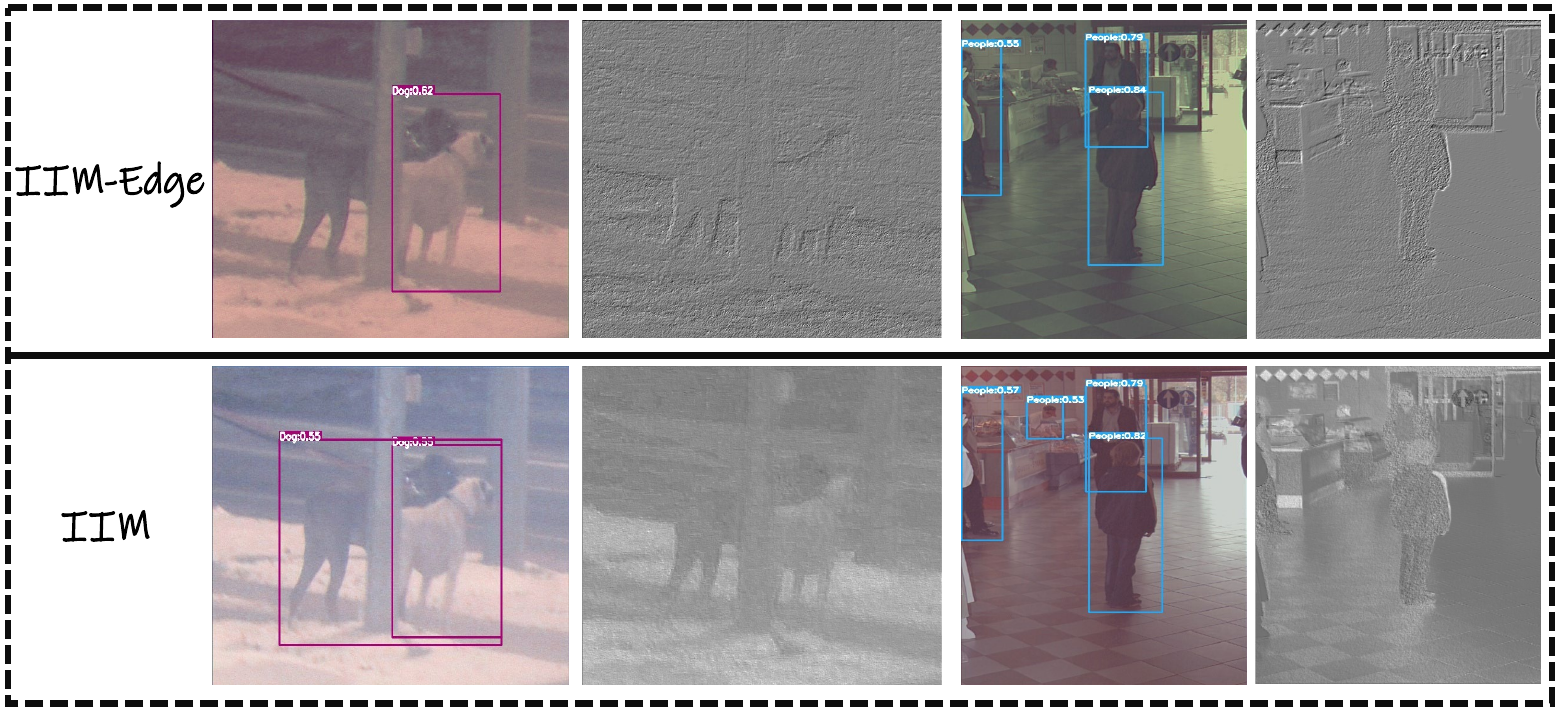}
   \caption{Visualization of the features (columns 2 and 4) generated by IIM-Edge and IIM(kernels are normalized for better visibility, we average the features across the channel dimensions and then conduct spatial normalization), along with detection results (columns 1 and 3). Best viewed by zooming in.}
\label{fig:feature_visualize}

\end{figure}

\noindent\textbf{Visualization:} Illumination-invariant features exhibit considerable diversity, but the diversity captured by fixed kernels is limited. We visualize and compare the fixed kernel and learnable kernel as shown in Fig.~\ref{fig:feature_visualize}. The features yielded by fixed kernels appear relatively uniform, primarily consisting of simple edge features. In contrast, learnable kernels extract more diverse patterns, resulting in visually richer and more informative representations.
\vspace{-2mm}

% \begin{table}[t]
% \centering
% \begin{tabular}{c|lll}
% \toprule
% Method    & AP$_{50}$ & AP$_{75}$ & mAP  \\ \midrule
% YOLOv3    & 56.3 & 34.8 & 33.4 \\
% + YOLA    & \textbf{56.8(+0.5)} & \textbf{35.7(+0.9)} & \textbf{33.9(+0.5)} \\ \midrule
% % RetinaNet & 55.1 & 38.6 & 36.1 \\
% % +yola     &  55.6(+0.5) & 39.0(+0.4) & 36.5(+0.4) \\ \hline
% TOOD      & 59.0 & 45.3 & 41.7 \\
% + YOLA     & \textbf{59.4(+0.4)} & \textbf{46.0(+0.7)} & \textbf{42.3(+0.6)}  \\ \bottomrule
% \end{tabular}

% \caption{Ablation study for YOLA on COCO {\ttfamily 2017val}.}
% \label{tab:coco2017}
% % \vspace{-5mm}
% \end{table}

\subsection{Generalization}
In this section, we broaden the application of the YOLA to the general object detection dataset COCO 2017, investigating the YOLA’s generalization capability beyond low-light object detection.
The metrics mAP (average for IoU [0.5:0.05:0.95]), AP$_{50}$, and AP$_{75}$ are adopted to evaluate performance on COCO {\ttfamily 2017val} (also called {\ttfamily minival}) as presented in Table~\ref{tab:coco2017}.
Specifically, we trained 12 epochs with 8 GPUs and a mini-batch of 1 per GPU in an initial learning rate of $1e$-$2$ by the SGD optimizer on both well-lit and over-lit (generated by brightening the origin image) scenarios.
By observing Table~\ref{tab:coco2017}, 
we can see that detectors integrated with YOLA in both scenarios exhibit notable improvements in performance.

\section{Conclusion}
\label{sec:conclusion}
In this work, we have revisited the complex challenge of object detection in low-light conditions and demonstrated the effectiveness of illumination-invariant features in improving detection accuracy in such environments. Our key innovation, the Illumination-Invariant Module (IIM), harnesses these features to great effect. By integrating a zero-mean constraint within the framework, we have effectively learned a diverse set of kernels. These kernels are adept at extracting illumination-invariant features, significantly enhancing detection precision. We believe that our developed IIM module can be instrumental in advancing low-light object detection tasks in future applications.
\paragraph{\textbf{Acknowledgement}:} 
This work was supported in part by National Natural Science Foundation of
China (NSFC) under grant No.62372091 and Natural Science Foundation of Sichuan Province under grant Nos. 2023NSFSC0462 and 2023NSFSC1972.

\bibliographystyle{plain}
\bibliography{final}

\appendix

\section{Appendix / supplemental material}
% \section{Derivation of IIM}
\label{sec:appendix}
\paragraph{\textbf{Derivation of IIM}.} 
Referring to the Eq.~\ref{equ:lccr} in the main text, the IIM defines a feature extracted from neighboring pixels. Consider a convolutional kernel $\mathcal{W}^{k \times k}$, where $k$ represents the kernel size. Here, $p_{i}$ and $w_{i}$ denote a pixel position and its associated weight within the kernel $\mathcal{W}$, with $i$ ranging from 1 to $k^{2}$.
\begin{equation}
    M_{rb} = \prod \limits_{i=1}^{k^2-1} \left(\frac{R_{p_{i}}}{B_{p_{i}}}\right)^{w_i}\left(\frac{B_{p_{i+1}}}{R_{p_{i+1}}}\right)^{w_{i+1}} = \prod \limits_{i=1}^{k^2} \left(\frac{R_{p_{i}}}{B_{p_{i}}}\right)^{w_i}
    \label{equ:lccr_}
\end{equation}

\begin{equation}
\begin{split}
    log(M_{rb}) &= \sum_{i=1}^{k} w_{i }log(R_{p_{i}}) -  \sum_{i=1}^{k} w_{i }log(B_{p_{i}}) \\
    &= \sum_{i=1}^{k} w_{i } (log(m(\Vec{n_{p_{i}}}, \Vec{l_{p_{i}}})) + log(e^{R_{p_{i}}}(\lambda)) + log(\rho^{R_{p_{i}}}(\lambda)) ) \\
    &- \sum_{i=1}^{k} w_{i } (log(m(\Vec{n_{p_{i}}}, \Vec{l_{p_{i}}})) + log(e^{B_{p_{i}}}(\lambda)) + log(\rho^{B_{p_{i}}}(\lambda)) ) \\
    &= \sum_{i=1}^{k} w_{i }log(e^{R_{p_{i}}}(\lambda)) - \sum_{i=1}^{k} w_{i }log(e^{B_{p_{i}}}(\lambda)) \\
    &+\sum_{i=1}^{k} w_{i }log(\rho^{R_{p_{i}}}(\lambda)) - \sum_{i=1}^{k} w_{i }log(\rho^{B_{p_{i}}}(\lambda))
\end{split}
\label{equ:iim_detail_}
\end{equation}

To eliminate the $e$ term, it is imperative to adhere to the following constraints::
\begin{equation}
\begin{split}
    \begin{cases}
    \sum_{i}^{k^2} w_{i}log(e^{R_{p_{i}}}(\lambda)) = 0 \\
    \sum_{i}^{k^2} w_{i}log(e^{B_{p_{i}}}(\lambda)) = 0
    \end{cases}
\end{split}
\label{equ:lccr_simple_cons_}
\end{equation}

Assuming that all pixels in a given convolutional kernel are neighboring pixels, we obtain $e^{R_{p_{i}}} \approx e^{R_{p_{j}}}$, where $i,j = 1, 2, \cdots, k^{2}, j \neq i$. The above constraints can be equivalently expressed as $\sum_{i}^{k^2} w_{i}=0$

\paragraph{\textbf{Illumination Invariant Loss}.} 
\label{part:IILoss}
As mentioned in Sec.~\ref{sec:lean_kernel}, to optimally constrain the kernel learning process and harness the full potential of illumination-invariant information, we further employ a consistency loss, denoted as Illumination Invariant Loss (II Loss). This loss function is specifically designed to align features extracted from pairs of images taken under different lighting conditions. The fundamental concept of the II Loss is to guarantee consistency in the features extracted from these images, regardless of the variations in illumination. This is achieved by leveraging a luminance transformation function $\sigma$ to adjust the illuminations,  as defined as follows:
\begin{align}
\mathclap{
    %\resizebox{.5\linewidth}{!}{$
        L = \begin{cases}\frac{1}{2}(f_{\mathcal{W}_{i}}( I) - f_{\mathcal{W}_{i}}( \sigma(I)))^2 & |f_{\mathcal{W}_{i}}( I) - f_{\mathcal{W}_{i}}( \sigma(I))| \leq \beta \\ |f_{\mathcal{W}_{i}}( I) - f_{\mathcal{W}_{i}}( \sigma(I))|-\frac{1}{2} \beta, & \text { otherwise. }\end{cases} %$ }
        }
\end{align}
In our experiments, we use the gamma transformations as the function for the function $\sigma$, setting $\beta$ empirically to 1, and scaling the II Loss to 0.01 of the other losses.

% \begin{figure}[t]
%   \centering
   
%   \includegraphics[width=.5\linewidth]{figs/kernel_v002.pdf}

%    \caption{Illustration of a trivial case (a), and visualization of performing $3\times3$ mean filtering on the kernel weights guided by with (b) and without (c) II loss.}
%    \label{fig:kernel}
%    % \vspace{-5mm}
% \end{figure}
\begin{figure}[t]
\vspace{10mm}
\begin{minipage}[t]{0.45\linewidth}
    \centering
    \includegraphics[width=1.\linewidth]{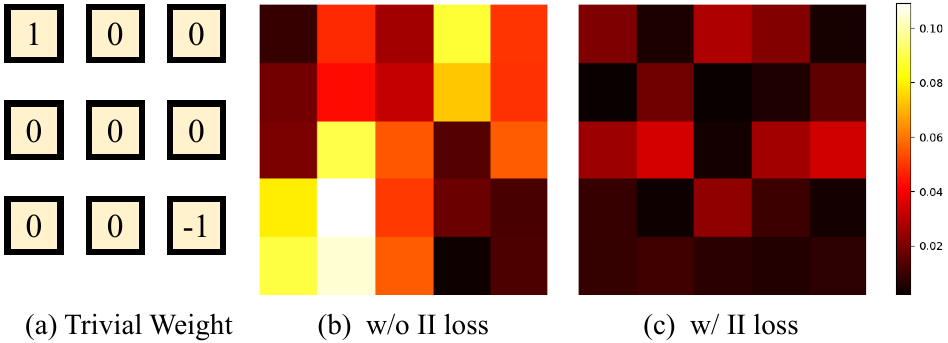}
    \caption{Illustration of a trivial case (a), and visualization of performing $3\times3$ mean filtering on the kernel weights guided by with (b) and without (c) II loss.}
    \label{fig:kernel}
\end{minipage}
\begin{minipage}[t]{.5\linewidth}
\vspace{-22mm}
\centering
\begin{adjustbox}{width=1\linewidth}
\begin{tabular}{l|c|ccc|cc}
\toprule
1) &  Dataset & IIM & II-Loss & $\mathcal{K}_{s}$ &  YOLOv3 & TOOD \\ \hline
   % &      &               &                  &     &    \\\hline
2)                            & \multirow{5}{*}{Exdark}     &               &                  &  3    & 71.0   & 72.5 \\
3)                            &                             &  \checkmark   &                  &  3    & 71.1  & 74.8 \\
4)                            &                             &  \checkmark   &   \checkmark     &  3    & \textbf{72.7}   & 75.0 \\ 
5)                            &                             &  \checkmark   &                  &  5    & 71.5  & 75.0 \\
6)                            &                             &  \checkmark  &   \checkmark      &  5    & \textbf{72.7} & \textbf{75.2}\\ \cline{2-7}
7)                            & \multirow{5}{*}{DarkFace}   &               &                  &  3    & 60.0  & 62.1 \\ 
8)                           &                             &  \checkmark   &                  &  3     & 61.0  & 66.9 \\  
9)                           &                             &  \checkmark   &   \checkmark     &  3     & \textbf{61.5}  & \textbf{67.4} \\
10)                           &                             &  \checkmark   &                  &  5    & 60.2   & 65.8 \\
11)                           &                             &   \checkmark  &   \checkmark     &  5    & 60.7  & 67.1 \\ \bottomrule
\end{tabular}
\end{adjustbox}
\caption{Ablation study of YOLOv3-based and TOOD-based detectors on ExDark and $UG^{2}+$DARK FACE datasets, where $\mathcal{K}_{s}$ denotes the kernel size within IIM.}
\label{tab:ablation}
\vspace{-0.3cm}
\end{minipage}

\end{figure}

As discussed in Sec.~\ref{sec:IIF}, we obtain illumination-invariant features by assuming neighboring pixels exhibit high similarity of illumination. Specifically, illumination items can be factored out by performing the subtraction among the neighboring pixels, which is accomplished by imposing the zero mean constraint on the convolutional kernels. However, to eliminate the illumination term ideally, it is necessary for the average value of adjacent positions within the kernel to approach zero. The sole constraint of a zero mean does not guarantee that the illumination elimination occurs strictly between adjacent pixels; it can occur between distant pixels as well.
For instance, Fig.~\ref{fig:kernel}(a) presents an example of a convolutional kernel that satisfies the zero-mean constraint. Even though this kernel has a zero mean, it fails to extract illumination-invariant features due to the relatively large spatial separation between the positive and negative positions. Unfortunately, as the convolutional kernel size increases, this issue becomes more pronounced and leads to a degradation in performance. 
To this end, the II Loss is proposed to encourage consistency of outputs from IIM across images with different illuminations, preventing trivial solutions within the kernel implicitly. As shown in Fig.~\ref{fig:kernel}(b)(c), we visualize the $5\times5$ kernel with and without the II Loss. For each position in the $5\times5$ kernel, we compute the mean value using a sliding window of size $3\times3$. It can be observed that without the II Loss, the local means within the kernel do not tend towards zero, indicating that the features extracted using this kernel are may not illumination-invariant. In contrast, when the II Loss is applied, the local values of the kernel are significantly constrained. To further validate the effectiveness of the II Loss, we present the results of the II Loss on different convolutional kernel sizes within IIM in rows 3$\sim$6 and 8$\sim$11 of Table~\ref{tab:ablation}. By comparing the performance gains of different convolutional kernel sizes, we can see that the larger kernel sizes lead to more significant improvements. This strongly suggests the effectiveness of our II Loss in constraining the degrees of freedom in the kernel.
\begin{table*}[t]
\resizebox{\textwidth}{!}{
\begin{tabular}{l|cccccccccccc|c}
    \toprule
\centering
Method   & Bicycle & Boat & Bottle & Bus  & Car  & Cat  &Chair  & Cup  & Dog  & Motorbike & People & Table & mAP$_{50}$ \\\midrule
Baseline & 79.8    & 72.1 & 70.9   & 82.8 & 79.5 & 64.4 &67.6   & 70.6 & 79.5 & 62.4      & 77.7   & 44.2  & 71.0    \\ 

MBLLEN~\cite{lv2018mbllen}                    & 77.5             & 72.5 & 70.2   & 80.7 & 80.6 & 65.0 &65.2   & 70.6 & 77.9 & 64.9      & 77.3   & 41.8  & 70.3    \\
KIND~\cite{zhang2019kindling}                 & 80.2             & 74.4 & 71.5   & 81.0 & 80.3 & 62.2 &61.3   & 67.5 & 75.8 & 62.1      & 75.9   & 40.9  & 69.4     \\
Zero-DCE~\cite{guo2020zero}                   & 81.8             & 74.6 & 70.1   & 86.3 & 79.5 & 61.0 &66.2   & 71.7 & 78.4 & 62.9      & 77.3   & 43.1  & 71.1     \\
EnlightenGAN~\cite{jiang2021enlightengan}     & 81.1             & 74.2 & 69.8   & 83.3 & 78.3 & 63.3 &65.5   & 69.3 & 75.3 & 62.5      & 76.7   & 41.0  & 70.0     \\
RUAS~\cite{liu2021retinex}                    & 76.4             & 69.2 & 62.7   & 77.3 & 74.9 & 59.0 &64.3   & 64.8 & 73.1 & 55.8      & 71.5   & 38.8  & 65.7    \\
SCI~\cite{ma2022toward}                       & 80.3             & 74.2 & 73.6   & 82.8 & 78.4 & 64.4 &65.8   & 71.3 & 78.1 & 62.7      & 78.2   & 42.4  & 71.0    \\
NeRCo~\cite{yang2023implicit}                 & 80.8             & 73.6 & 66.3   & 81.3 & 75.6 & 62.8 &62.5   & 67.7 & 75.6 & 61.8      & 75.1   & 39.0  & 68.5    \\
SMG~\cite{xu2023low}                          & 78.1             & 72.1 & 65.8   & 81.6 & 78.3 & 63.7 &64.5   & 67.6 & 76.3 & 57.4      & 73.7   & 42.4  & 68.5    \\
DENet~\cite{qin2022denet}                     & 81.1             & 75.0 & 73.9   & 87.1 & 79.7 & 63.5 &66.3   & 69.6 & 76.3 & 61.4      & 76.7   & 44.9  & 71.3    \\
PENet~\cite{yin2023pe}                        & 76.5             & 71.9 & 67.4   & 84.2 & 78.0 & 59.9 &64.6   & 66.7 & 74.8 & 62.5      & 73.9   & 45.1  & 68.8    \\
MAET~\cite{cui2021multitask}                  & 81.5             & 73.7 & 74.0   & 88.2 & 80.9 & 68.8 &66.9   & 71.8 & 79.3 & 60.2      & 78.8   & 46.3  & 72.5     \\ \midrule
Ours                                          & 82.4    & 74.0 & 72.7   &85.4 & 81.0 & 67.2 &66.5   & 71.5 & 81.8 & 65.2      & 78.6   & 45.7  & 72.7   \\ \bottomrule
\end{tabular}
}
\caption{Quantitative comparisons of the ExDark dataset based on YOLOv3 detector.}
\label{tab:exdark_yolov3_detailed}
\end{table*}
\begin{table*}[t]
\resizebox{\textwidth}{!}{
\begin{tabular}{l|cccccccccccc|c}
    \toprule
\centering
Method   & Bicycle & Boat & Bottle & Bus  & Car  & Cat  &Chair  & Cup  & Dog  & Motorbike & People & Table & mAP$_{50}$ \\\midrule
Baseline                          & 80.6    & 75.8 & 71.1   & 88.1 & 76.8 & 70.4 &66.8   & 69.2 & 85.4 & 61.5      & 76.1   & 48.2  & 72.5    \\ 
MBLLEN~\cite{lv2018mbllen}        & 80.8    & 77.8 & 72.8   & 89.3 & 78.7 & 73.5 &67.5   & 69.4 & 85.2 & 62.9      & 77.3   & 47.2  & 73.5    \\
KIND~\cite{zhang2019kindling}     & 81.7    & 77.7 & 70.3   & 88.4 & 78.1 & 69.7 &67.2   & 67.8 & 84.1 & 61.6      & 76.6   & 47.8  & 72.6     \\
Zero-DCE~\cite{guo2020zero}       & 81.8    & 79.0 & 72.9   & 89.6 & 77.9 & 71.9 &68.5   & 69.8 & 84.8 & 62.9      & 78.0   & 49.5  & 73.9     \\
EnlightenGAN~\cite{jiang2021enlightengan}     & 80.7    & 77.6 & 70.4   & 88.8 & 76.9 & 70.6 &67.9   & 68.7 & 84.4 & 62.2      & 77.5   & 49.6  & 73.0     \\
RUAS~\cite{liu2021retinex}        & 78.4    & 74.3 & 67.4   & 85.1 & 72.4 & 67.7 &67.3   & 65.2 & 77.9 & 56.1      & 73.4   & 47.0  & 69.4    \\
SCI~\cite{ma2022toward}           & 81.3    & 78.1 & 71.6   & 89.4 & 77.6 & 71.1 &68.0   & 70.9 & 85.0 & 63.0      & 77.2   & 49.2  & 73.5     \\
NeRCo~\cite{yang2023implicit}     & 78.8    & 75.6 & 70.8   & 87.6 & 75.7 & 69.1 &66.8   & 69.5 & 82.5 & 59.9      & 76.0   & 49.3  & 71.8     \\
SMG~\cite{xu2023low}              & 78.2    & 75.9 & 69.9   & 87.3 & 75.1 & 71.3 &66.5   & 67.2 & 84.2 & 60.1      & 75.1   & 46.7  & 71.5     \\
DENet~\cite{qin2022denet}         & 80.9    & 78.2 & 70.9   & 88.3 & 77.5 & 71.6 &67.2   & 70.3 & 87.3 & 62.0      & 77.3   & 49.9  & 73.5    \\
PENet~\cite{yin2023pe}            & 76.0    & 72.3 & 66.7   & 84.4 & 72.2 & 65.4 &63.3   & 65.8 & 79.1 & 53.1      & 71.0   & 44.6  & 67.8    \\
MAET~\cite{cui2021multitask}      & 80.5    & 77.3 & 74.0   & 90.1 & 78.3 & 73.4 &69.6   & 70.7 & 86.6 & 64.4      & 77.6   & 48.5  & 74.3     \\ \midrule
Ours                              & 83.9    & 78.7 & 75.3   & 88.8 & 79.0 & 73.4 &69.9   & 71.9 & 86.8 & 66.3      & 78.3   & 49.8  & 75.2   \\ \bottomrule
\end{tabular}
}
\caption{Quantitative comparisons of the ExDark dataset based on TOOD detector.}
\label{tab:exdark_tood_detailed}
\end{table*}
\paragraph{\textbf{Detailed Results on ExDark}.} 
In this section, we report the average precision for each category of the ExDark dataset as shown in Table~\ref{tab:exdark_yolov3_detailed} and ~\ref{tab:exdark_tood_detailed}.
Note that, we further introduce more advanced LLIE methods, including Zero-DCE~\cite{guo2020zero}, EnlightenGAN~\cite{jiang2021enlightengan}, SCI~\cite{ma2022toward}, and NeRCo~\cite{yang2023implicit} based on YOLOv3 and TOOD. Unfortunately, despite the outstanding performance exhibited by these LLIE methods in image restoration tasks,  they struggle with effectively enhancing specific downstream tasks. For example, the state-of-the-art LLIE method, NeRCo,  exhibits the worst performance compared to other LLIE methods. This phenomenon further proves the existence of the gap between optimization goals for image restoration and object detection tasks. Additionally,  compared to end-to-end approaches, such cascade paradigms limit the potential for deploying these LLIE-based low-light detection techniques to practical applications.

\paragraph{\textbf{YOLA vs. FeatEnHancer}.} 
% \section{YOLA vs. FeatEnHancer}
For a fair comparison, we follow the FeatEnHancer’s~\cite{hashmi2023featenhancer} experimental setting to implement the RetinaNet~\cite{lin2017focal}-based detectors as shown in Table~\ref{tab:featenhancer}. We can see that even though our baseline implementation on the ExDark dataset is inferior to  FeatEnHancer’s, the integration of YOLA enables our method to achieve the best performance (1.9~mAP significant improvement compared to baseline). For $UG^{2}+$DARK FACE dataset,  
FeatEnHancer decreases the baseline performance by 0.1~mAP, which is attributed to hierarchical features that failed to be captured by RetinaNet, as claimed in~\cite{hashmi2023featenhancer}. In contrast, our YOLA, triggered from the physics-based model perspective without elaborate design, surpassing the baseline with a remarkable improvement of 2.5~mAP. It strongly suggests the generalizability and effectiveness of YOLA.
\begin{figure*}[t]
  \centering
  \includegraphics[width=1.\linewidth]{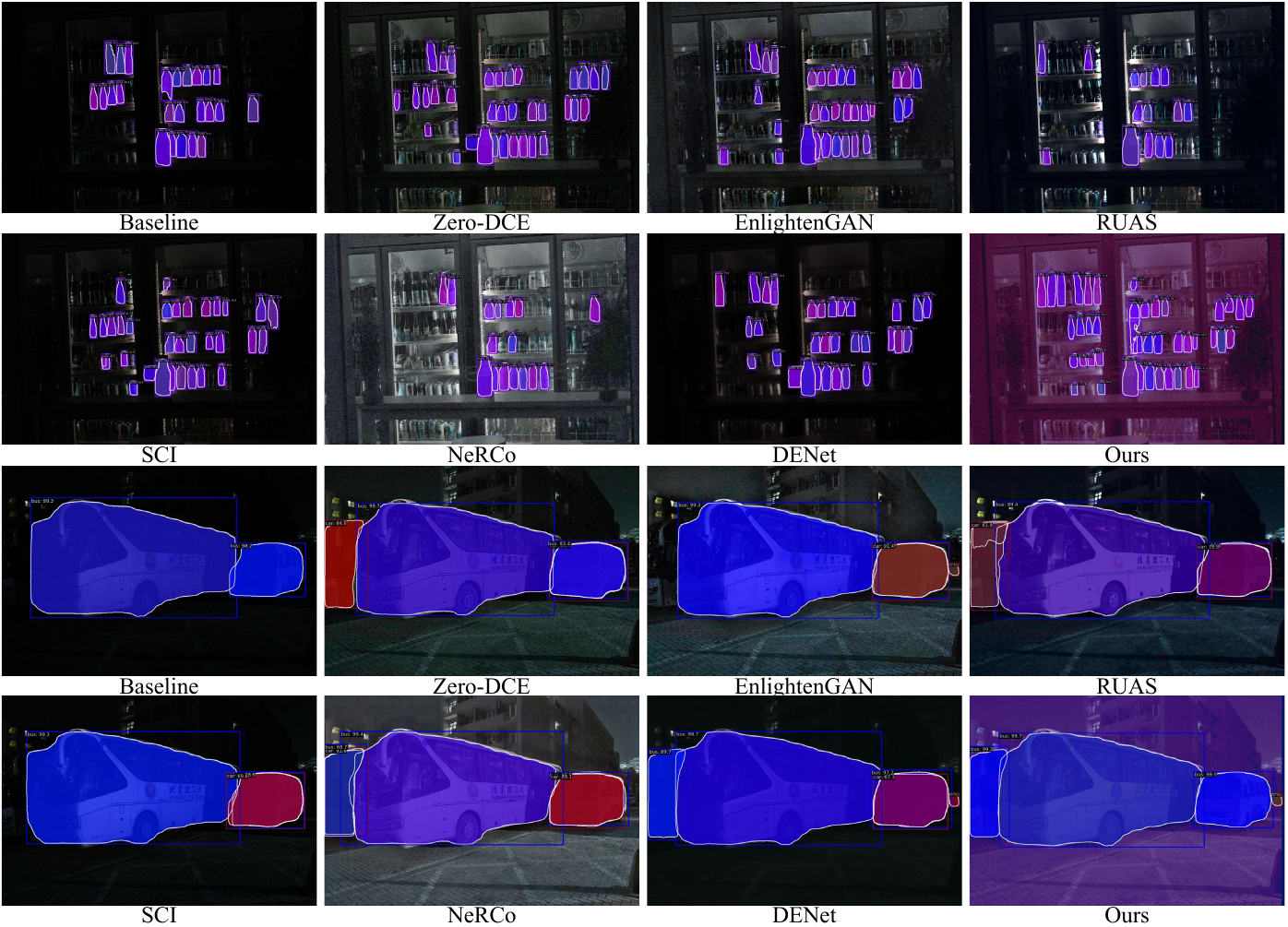}

   \caption{Qualitative comparisons of Mask R-CNN-based detector on LIS dataset. Our YOLA shows more comprehensive segmentation and detection results, with an increased number of bottles detected (top 2 rows) and successful recognition of the challenging car (right side of the bottom 2 rows). Best viewed with zooming in.}
   \label{fig:instance1}
\end{figure*}

\begin{table*}[t]
\begin{minipage}[t]{0.48\linewidth}
\centering
\resizebox{0.9\linewidth}{!}{
\begin{tabular}{c|l|l}
\toprule
Dataset                 &Method                             & mAP$_{50}$    \\ \midrule
\multirow{4}{*}{Exdark} &Baseline                           &72.1            \\
                        &w/ FeatEnHancer~\cite{hashmi2023featenhancer}    &72.6(+\textcolor{red}{\textbf{0.5}})  \\  \cline{2-3}
                        &Baseline$^{\dagger}$               &70.9           \\
                        &w/ YOLA                            &\textbf{72.8(+\textcolor{red}{1.9})}   \\  \midrule
\multirow{4}{*}{DarkFace} &  Baseline & 47.3 \\
                        & w/ FeatEnHancer~\cite{hashmi2023featenhancer} &47.2(-\textcolor{blue}{\textbf{0.1}}) \\ \cline{2-3}
                        &Baseline$^{\dagger}$               &50.2    \\  
                        &w/ YOLA                            &\textbf{52.7(+\textcolor{red}{2.5})}     \\ \bottomrule
\end{tabular}
}
\caption{Quantitative comparisons (YOLA vs. FeatEnHancer) of ExDark and $UG^{2}+$DARK FACE datasets based on RetinaNet. \textcolor{red}{Red} and \textcolor{blue}{blue} colors represent \textbf{improvement} and \textbf{degradation} of performance, respectively, compared to the baseline. $\dagger$ indicates our implemented baseline.} 
\label{tab:featenhancer}
\end{minipage}
\begin{minipage}[t]{0.5\linewidth}
\resizebox{1.0\linewidth}{!}{
\begin{tabular}{l|ccc|ccc}
\toprule
Method        &AP$^{seg}$          &AP$^{seg}_{50}$       &AP$^{seg}_{75}$      &AP$^{box}$              &AP$^{box}_{50}$     &AP$^{box}_{75}$         \\ \midrule
Baseline      & 34.2                 & 55.6                & 34.7                &41.3                     &63.9                       &44.6            \\ 
DENet~\cite{qin2022denet}         & 38.6                 & 61.7                & 39.8                &46.4                     &70.1                       &51.0            \\ 
PENet~\cite{yin2023pe}         & 36.1                 & 58.8                & 36.4                &43.6                     &67.3                       &47.1            \\ 
Zero-DCE~\cite{guo2020zero}      & 38.7                 & 62.0                & 39.0                &46.4                     &70.0                       &50.9            \\ 
EnlightenGAN~\cite{jiang2021enlightengan}  & 38.4                 & 61.5                & 39.2                &45.8                     &69.5                       &49.7            \\ 
RUAS~\cite{liu2021retinex}        & 36.1                 & 58.6                & 36.4                &43.8                     &66.7                       &48.0            \\ 
SCI~\cite{ma2022toward}           & 36.5                 & 59.5                & 37.0                &44.3                     &67.3                       &48.4            \\
NeRCo~\cite{yang2023implicit}     & 36.7                 & 60.3                & 38.6                &44.6                     &68.3                       &48.6           \\
SMG~\cite{xu2023low}              & 37.4                 & 60.3                & 38.7                &44.7                     &67.4                       &49.2   
\\ \midrule
Ours          &\textbf{39.8}         &\textbf{63.5}        &\textbf{41.4}       &\textbf{47.5}            &\textbf{70.9}              &\textbf{51.8}             \\ \bottomrule
\end{tabular}
}
\caption{Quantitative comparisons of the LIS dataset based on Mask RCNN, where AP$^{seg}$ and AP$^{box}$ indicate the average precision of segmentation and detection, respectively.}
\label{tab:instance}
\end{minipage}
%\vspace{-5mm}
\end{table*}
\begin{figure*}[t]
  \centering
  \includegraphics[width=1.\linewidth]{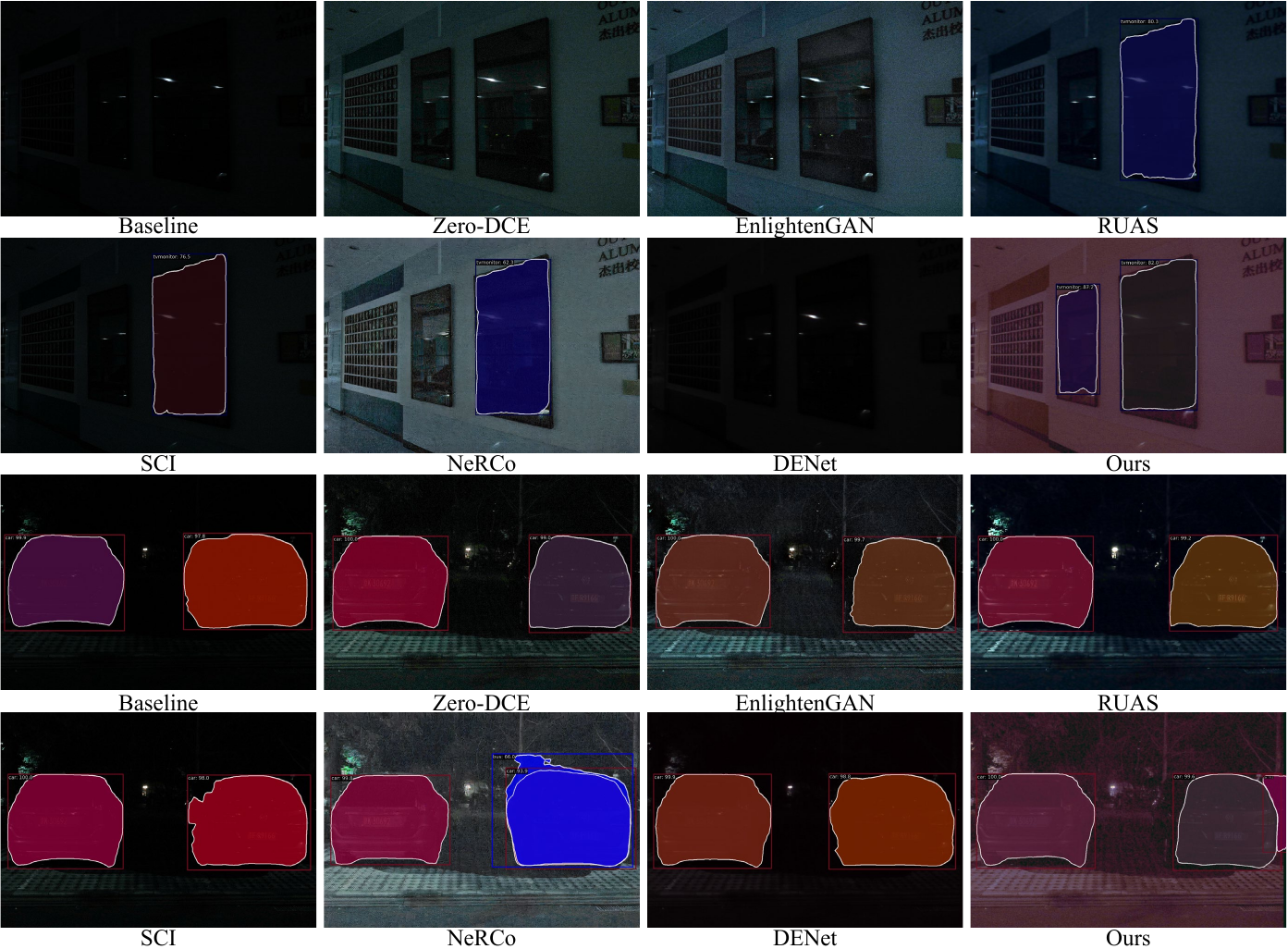}
   \caption{Qualitative comparisons of Mask R-CNN-based detector on LIS dataset. Our YOLA outperforms LLIE-based and low-light object detection methods. Best viewed with zooming in.}
   \label{fig:instance2}
\end{figure*}
\paragraph{\textbf{More Visualization}.} 
In Fig.~\ref{fig:instance1} and ~\ref{fig:instance2}, additional visual results are presented, showcasing selected challenging cases. In comparison to other methods, our method exhibits superior recall and more precise segmentation performance under extreme low-light conditions.
\paragraph{\textbf{YOLA on Lowl-light Instance Segmentation}.} 
To further explore YOLA's capabilities, we also evaluate YOLA in the low-light instance segmentation tasks. We report the quantitative comparisons of several advanced LLIE and low-light object methods using Mask R-CNN~\cite{he2017mask} on the low-light instance segmentation (LIS)~\cite{chen2023instance} dataset, as shown in Table~\ref{tab:instance}.
We can see that our YOLA achieves the best performance across all metrics,  indicating that YOLA not only facilitates low-light object detection but also low-light instance segmentation.

\end{document}